\documentclass[conference]{IEEEtran}
\IEEEoverridecommandlockouts
\usepackage{cite}
\usepackage{amsmath,amssymb,amsfonts}
\usepackage{algorithmic}
\usepackage{graphicx}
\usepackage{textcomp}
\usepackage{xcolor}
\usepackage{url}
 \usepackage{booktabs}
\def\BibTeX{{\rm B\kern-.05em{\sc i\kern-.025em b}\kern-.08em
    T\kern-.1667em\lower.7ex\hbox{E}\kern-.125emX}}
\begin{document}

\title{An Edge AI Solution for Space Object Detection\\
\thanks{We acknowledge the support of the Natural Sciences and Engineering Research Council of Canada (NSERC), [funding reference number RGPIN-2022-03364]}
}

\author{\IEEEauthorblockN{
Wenxuan Zhang\IEEEauthorrefmark{1} and
Peng Hu\IEEEauthorrefmark{1}\IEEEauthorrefmark{2}}
\IEEEauthorblockA{\IEEEauthorrefmark{1}Faculty of Mathematics, University of Waterloo, Waterloo, Canada}
\IEEEauthorblockA{\IEEEauthorrefmark{2}
Dept. of Electrical and Computer Engineering, University of Manitoba, Winnipeg, Canada}
{v39zhang@uwaterloo.ca, peng.hu@umanitoba.ca}
% <-this % stops an unwanted space
}

\maketitle

\begin{abstract}
Effective Edge AI for space object detection (SOD) tasks that can facilitate real-time collision assessment and avoidance is essential with the increasing space assets in near-Earth orbits. In SOD, low Earth orbit (LEO) satellites must detect other objects with high precision and minimal delay. We explore an Edge AI solution based on deep-learning-based vision sensing for SOD tasks and propose a deep learning model based on Squeeze-and-Excitation (SE) layers, Vision Transformers (ViT), and YOLOv9 framework. We evaluate the performance of these models across various realistic SOD scenarios, demonstrating their ability to detect multiple satellites with high accuracy and very low latency.
\end{abstract}


\section{Introduction}
As low-Earth orbit (LEO) satellite constellations expand, they enable global Internet access and serve as relay systems for deep-space missions, making space assets increasingly critical. However, ensuring the safety and sustainability of thousands of LEO satellites is a growing challenge. Rising collision risks with other space objects generate debris of various sizes, threatening spacecraft operations and the overall space environment. Traditional sensing technologies such as LiDAR and ground-based radar lack the high-precision, low-power, and low-latency capabilities required for effective SOD. In contrast, onboard vision sensing offers a cost-effective, low-power alternative for real-time detection. YOLOv9, introduced by \cite{wang2024yolov9learningwantlearn}, leverages the Generalized Efficient Layer Aggregation Network (GELAN) and Programmable Gradient Information (PGI) to enhance detection performance while maintaining low computational complexity, making it well suited for SOD onboard. However, models based on the Convolutional Neural Network (CNN) often struggle with small object detection due to their limited contextual awareness, while Vision Transformers (ViTs) \cite{Dosovitskiy2020AnII} improve long-range dependency modeling but at the cost of higher computational demands. This study explores a hybrid deep learning framework that integrates CNNs, ViTs, and Squeeze-and-Excitation (SE) blocks to improve onboard SOD. We also introduce the SODv2 dataset, which simulates real-world LEO conditions, and evaluate our models against GELAN-t and GELAN-ViT.

\section{Model}
\begin{figure}
    \centering
    \includegraphics[width=\linewidth]{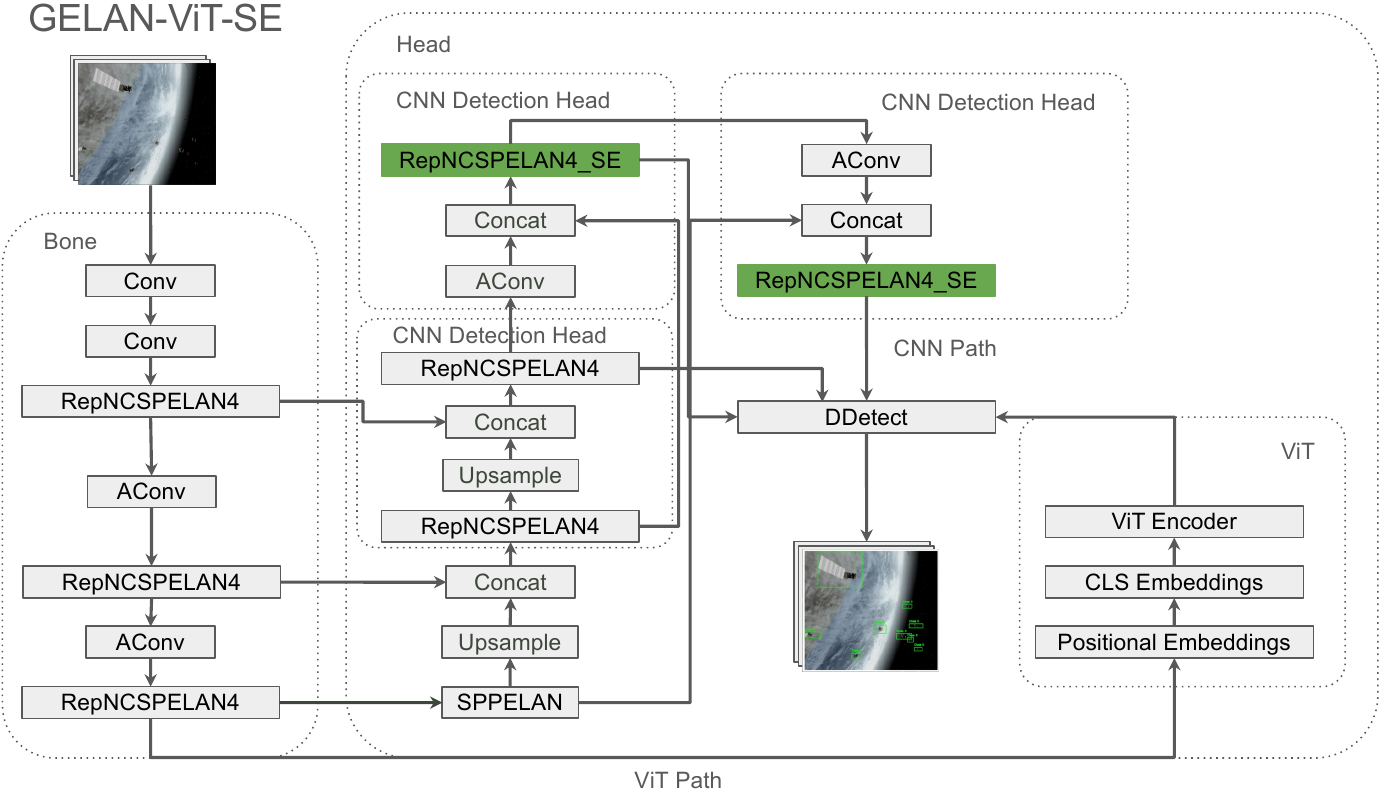}
    \caption{The architecture of the GELAN-ViT-SE model.}
    \label{fig:GELAN_ViT_SE}
    \vspace{-15pt}
\end{figure}

The proposed GELAN-ViT-SE model is built on a hybrid architecture that integrates CNNs and ViTs to improve feature extraction. It extends our previous GELAN-ViT design \cite{zhang2024sensing} by incorporating SE block to introduce channel-wise attention to CNN layers. The architecture of GELAN-ViT-SE is shown in Fig. \ref{fig:GELAN_ViT_SE}, where the last two detection heads replace the RepNCSPELAN4 module with RepNCSPELAN4\_SE, an enhanced version that integrates an SE block \cite{Liu2021AnI2} for improved channel-wise recalibration.

\begin{figure}
    \centering
    \includegraphics[width=0.92\linewidth]{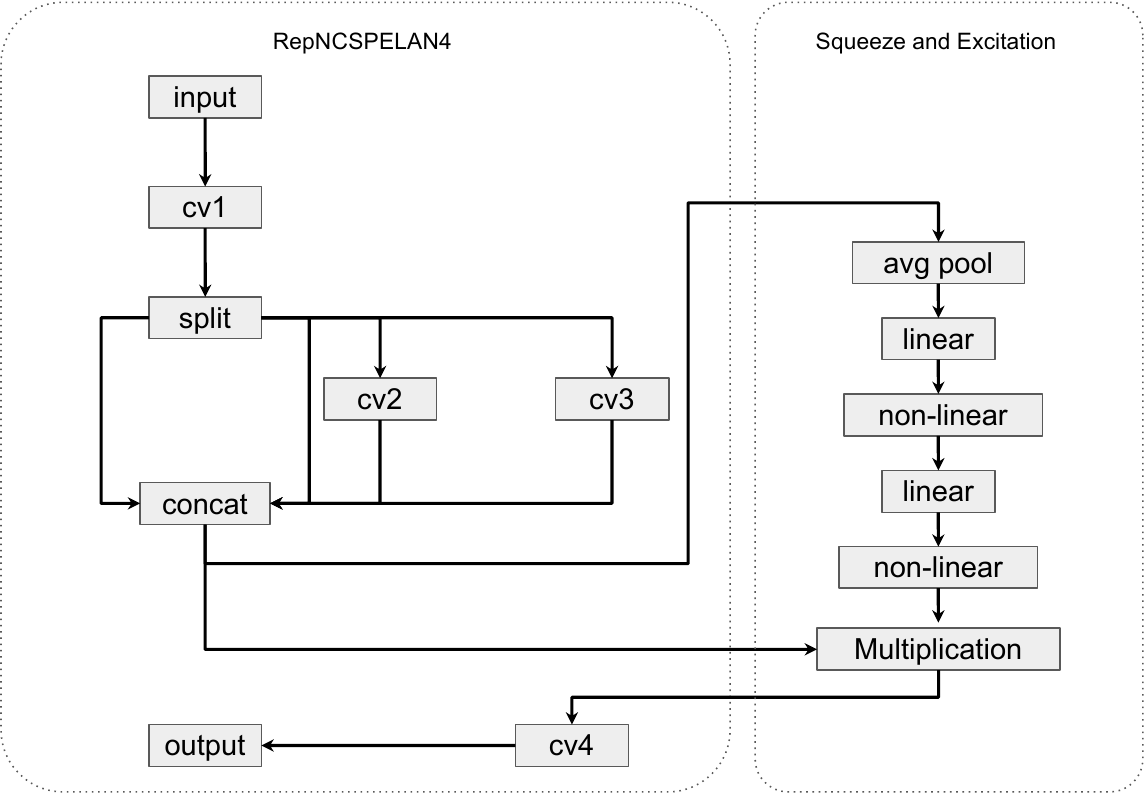}
    \caption{The architecture of the RepNCSPELAN4\_SE module. The SE layer is highlighted with a distinct background color.}
    \label{fig:SE}
\end{figure}

The architecture of RepNCSPELAN4\_SE is illustrated in Fig. \ref{fig:SE}. An initial convolutional layer \(cv1\) processes the input feature map \(x\) into \(x'\), which is split into \(x_0\) and \(x_1\). These are further processed by \(cv2\) and \(cv3\), generating \(x_0'\) and \(x_1'\). The resulting feature maps are concatenated into \(y\), which is then recalibrated by the SE block. Within the SE layer, a global average pooling computes a channel descriptor \(s_c\) for each channel, and \(s_c\) then passed through two fully connected layers to generate attention weights \(z_c\). These attention weights are applied element-wise to the original channel, resulting in a recalibrated output \( y' \). Finally, the recalibrated feature map \( y' \) are processed by the convolutional layer \(cv4\) to produce the output.

\section{Dataset}

\begin{figure}[ht]
    \centering
    \includegraphics[width=\linewidth]{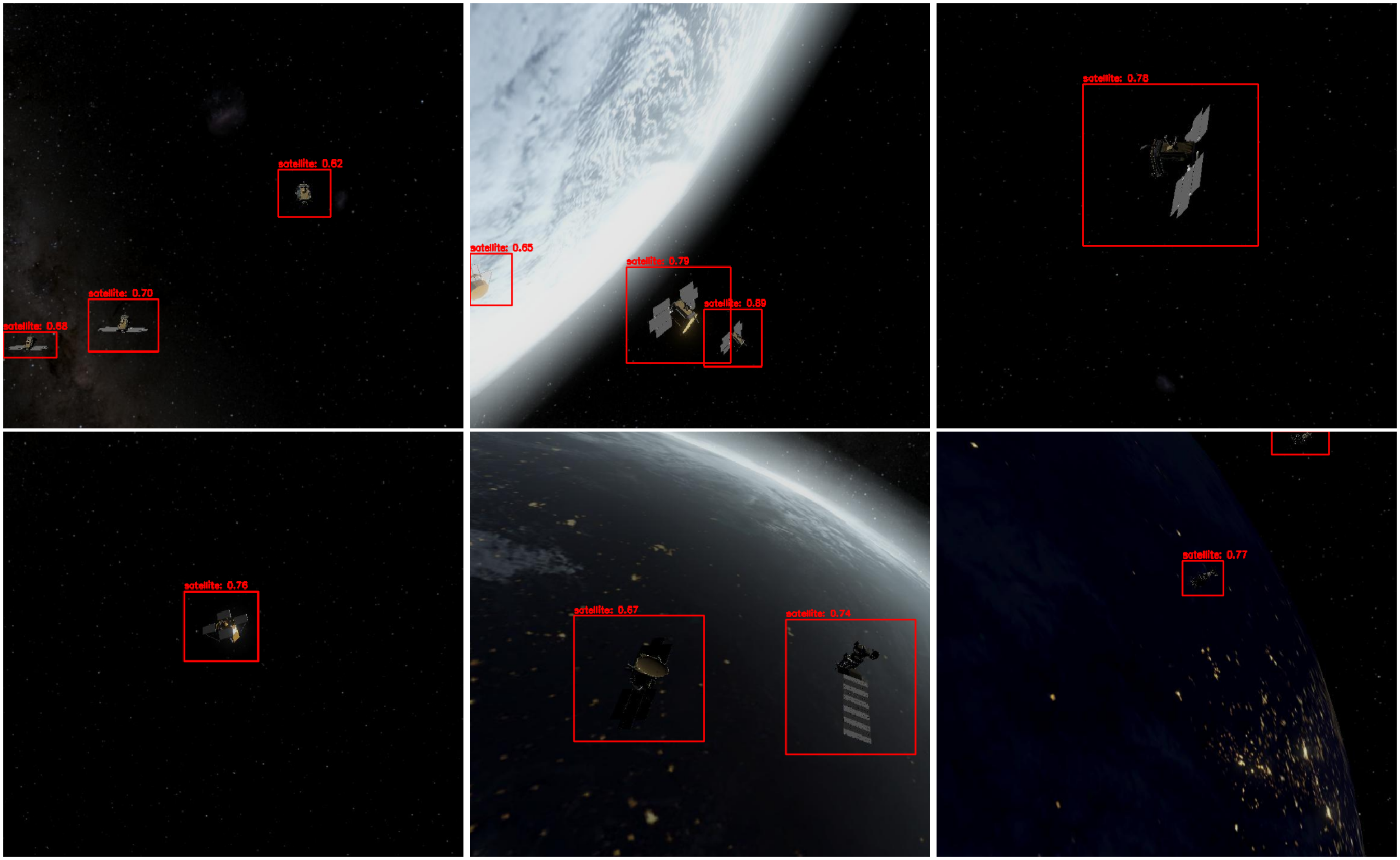}
    \caption{Sample SODv2 images with detection bounding boxes and confidence scores for satellite objects.}
    \label{fig:sodv2_examples}
    %\vspace{-12pt}
\end{figure}

We developed the SODv2 dataset\footnote{Available: \url{https://github.com/AEL-Lab/satellite-object-detection-dataset-v2}} to simulate LEO satellite orbits in realistic space environments. We modeled a dynamic solar system with accurate celestial scales and distances. LEO satellites were placed at altitudes between 500 km and 600 km with random orbital configurations. Each satellite was equipped with an onboard camera with a fixed 45-degree field of view (FoV), which sequentially captured images as the satellite moved through its orbit. The camera script automatically recorded distances and annotated bounding boxes for satellites detected within a 5 km range. Fig. \ref{fig:sodv2_examples} illustrates sample detection results, where each bounding box highlights a detected satellite along with the model’s confidence score generated by GELAN-ViT-SE.

\section{Performance Evaluation}

To evaluate our models on the SODv2 dataset, we trained each model on 450 images and tested it on a separate set of 150 images. Each model was trained three times, and we report the average performance across runs. Training was conducted on a high-performance computing cluster with a Tesla V100-SXM2-32GB GPU, while inference speed was tested on a Jetson Orin Nano to simulate onboard LEO satellite conditions. All models were trained for 1000 epochs with a batch size of 16 to ensure fair comparisons. For computational efficiency measurements, inference results were averaged over 25 consecutive runs, with the first five discarded to minimize variability due to resource allocation. To reflect real-time processing conditions, inference speed and power consumption were evaluated using a batch size of 1.

Table \ref{tab:results} presents the detection performance of different models across the specified distance range. Compared to GELAN-t, the GELAN-ViT framework improves satellite identification accuracy while reducing computational complexity, and GELAN-ViT-SE further enhances detection performance.

Table \ref{tab:jetson} provides the runtime analysis on the NVIDIA Jetson Orin Nano, highlighting the trade-offs between inference time, memory usage, and power consumption. While GELAN-t achieves the lowest latency, it consumes more power than ViT-based models. In contrast, GELAN-ViT-SE delivers stable inference performance with slightly higher memory usage.

\begin{table}[ht]
    \centering
    \caption{Results for 0 km to 5 km Distance Range}
    \resizebox{0.9\linewidth}{!}{
    \begin{tabular}{lcccc}
        \toprule
        \textbf{Model} & \textbf{mAP50} & \textbf{mAP50:95} & \textbf{GFLOPs} \\
        \midrule
        GELAN-t         & 0.721 & 0.266 & 7.3\\
        GELAN-ViT       & 0.737 & 0.265 & \textbf{5.6}\\
        GELAN-ViT-SE    & \textbf{0.751} & \textbf{0.274} & \textbf{5.6} \\
        \bottomrule
    \end{tabular}
    }
    \label{tab:results}
\end{table}

\begin{table}[ht!]
    \centering
    \caption{Evaluation on Jetson Orin Nano}
    \resizebox{\linewidth}{!}{
    \begin{tabular}{lccc}
        \toprule
        \textbf{Model Name} & \textbf{Inf. Time} & \textbf{Peak RAM} & \textbf{Peak Power} \\
        \midrule
        GELAN-t         & 46.14 ms & 2.377 GB & 2080.7 mW \\
        GELAN-ViT       & 56.47 ms & 2.463 GB & 1988.6 mW \\
        GELAN-ViT-SE    & 58.88 ms & 2.557 GB & 2028.7 mW \\
        \bottomrule
    \end{tabular}
    }
    \label{tab:jetson}
    %\vspace{-12pt}
\end{table}

\section{Conclusion}
In this paper, we introduced GELAN-ViT-SE model and the SODv2 dataset. Our experiments demonstrate the feasibility of the proposed Edge AI solution, showing that the addition of SE blocks enhances detection accuracy while maintaining similar computational efficiency. This confirms that channel-wise recalibration in the GELAN framework is a promising strategy for performance improvement. A future direction would be to expand the scope of experiments by testing on diverse datasets and exploring different hyperparameter configurations to analyze the advantages of each model.

\bibliographystyle{IEEEtran}
\bibliography{IEEEabrv,references}
\vspace{12pt}

\end{document}